\pdfoutput=1

\documentclass[11pt]{article}

\usepackage[preprint]{acl}

\usepackage{times}
\usepackage{latexsym}

\usepackage[T1]{fontenc}

\usepackage[utf8]{inputenc}

\usepackage{microtype}

\usepackage{inconsolata}

\usepackage{graphicx}

\usepackage{booktabs}
\usepackage{subcaption}
\usepackage{multirow}
\usepackage{wrapfig} 
\usepackage{colortbl}
\usepackage{enumitem}
\usepackage{kotex} 
\usepackage[most]{tcolorbox}
\usepackage{adjustbox}

\definecolor{Highlight}{HTML}{39b54a}  
\definecolor{Cerulean}{HTML}{00A2E3}  

\newcommand{\PAR}[1]{\vskip4pt \noindent {\bf #1.~}} 

\title{Harnessing PDF Data for Improving Japanese Large Multimodal Models}

\author{
Jeonghun Baek$^{\spadesuit}$\quad Akiko Aizawa$^{\diamondsuit}$\quad Kiyoharu Aizawa$^{\spadesuit}$ \\
$^{\spadesuit}$The University of Tokyo \quad
$^{\diamondsuit}$National Institute of Informatics \\
\texttt{baek@hal.t.u-tokyo.ac.jp}\\
\url{https://github.com/ku21fan/PDF-JLMM}
}

\begin{document}
\maketitle

\begin{abstract}
Large Multimodal Models (LMMs) have demonstrated strong performance in English, but their effectiveness in Japanese remains limited due to the lack of high-quality training data. Current Japanese LMMs often rely on translated English datasets, restricting their ability to capture Japan-specific cultural knowledge. To address this, we explore the potential of Japanese PDF data as a training resource, an area that remains largely underutilized. We introduce a fully automated pipeline that leverages pretrained models to extract image-text pairs from PDFs through layout analysis, OCR, and vision-language pairing, removing the need for manual annotation. Additionally, we construct instruction data from extracted image-text pairs to enrich the training data. To evaluate the effectiveness of PDF-derived data, we train Japanese LMMs and assess their performance on the Japanese LMM Benchmark. Our results demonstrate substantial improvements, with performance gains ranging from 2.1\% to 13.8\% on Heron-Bench. Further analysis highlights the impact of PDF-derived data on various factors, such as model size and language models, reinforcing its value as a multimodal resource for Japanese LMMs.
\end{abstract}

\begin{figure}[t]
  \includegraphics[width=\linewidth]{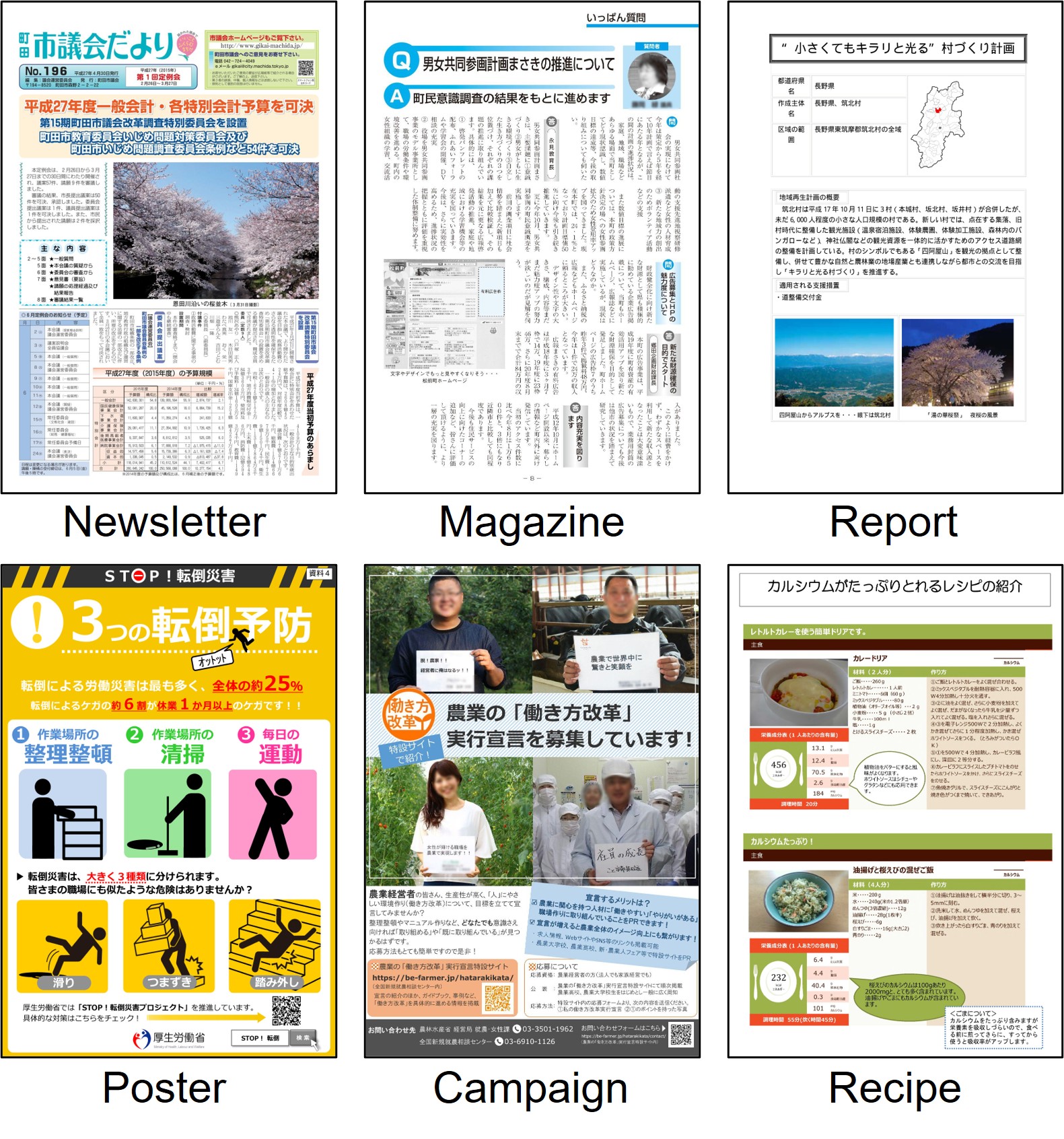}
  \vspace{-8mm}
  \caption{\textbf{Examples of PDF data.} 
  We use various types of PDF data for Japanese LMM training.
  }
  \label{fig:pdf-category}
\end{figure}

\begin{figure*}[t]
  \includegraphics[width=\linewidth]{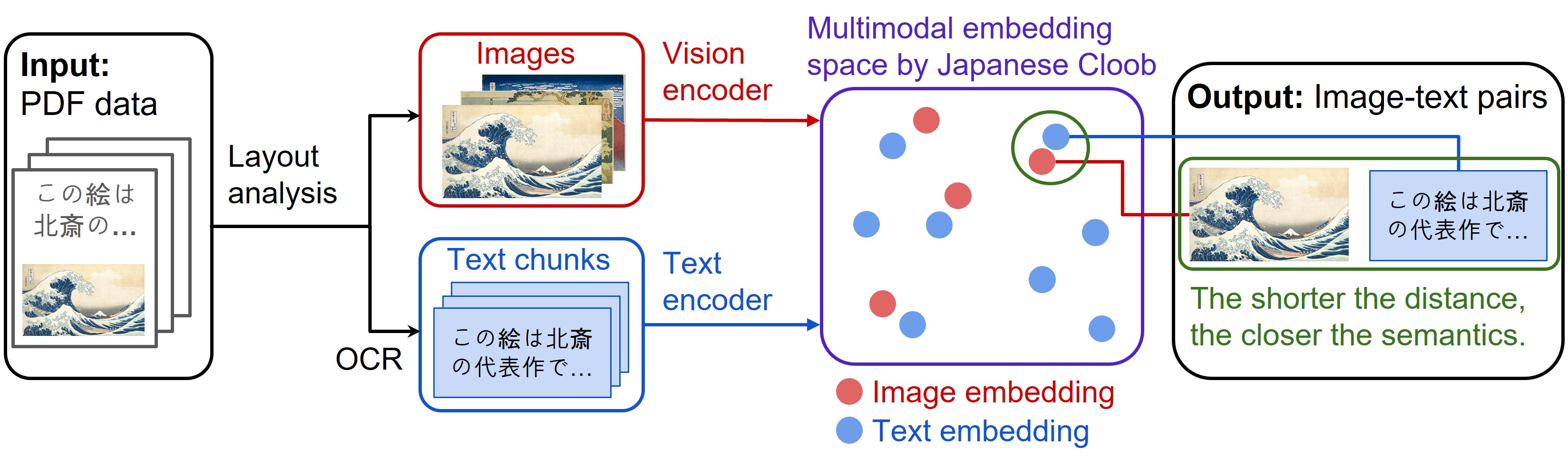}
  \vspace{-8mm}
  \caption{\textbf{An automated pipeline for extracting image-text pairs from PDFs.} 
  It leverages pretrained models for layout analysis, OCR, and vision-language pairing.}
  \label{fig:overview}
\end{figure*}

\section{Introduction}
Large Multimodal Models (LMMs) have achieved high performance in English~\cite{openai2024gpt4o,geminipro,llama3.1,yang2024qwen2}, and their development is now expanding to other languages. Recently, several open-source Japanese LMMs have been released~\cite{akiba2025evolutionary,BlipJapaneseStableLM,inoue2024heron,cyberagent2024llava,VILAjp}. While these models perform reasonably well, they still lag behind their English counterparts, partly due to the limited availability of Japanese training data.

Unlike English, where large-scale public image-text pairs exist, Japanese models often rely on translated English data~\cite{JapaneseInstructBLIPAlpha,JapaneseStableVLM,BlipJapaneseStableLM,inoue2024heron}. This means Japanese LMMs primarily learn content from western sources, lacking exposure to Japan-specific cultural knowledge.

To address this issue, we utilize Japanese PDF data to incorporate culturally relevant knowledge into LMM training. Unlike existing multimodal datasets, which are primarily web-based~\cite{coco,sharma2018conceptual,schuhmann2022laion}, PDFs contain a vast amount of valuable but underutilized information from books and documents. Despite this, research on integrating PDF data into LMM training remains limited. To our knowledge, no existing work has leveraged PDF data to enhance Japanese LMMs.

We investigate whether PDF data can effectively enhance Japanese LMMs. However, manually annotating large-scale PDFs is costly. To overcome this, we develop a fully automated pipeline, as shown in Figure~\ref{fig:overview}. 
It extracts image-text pairs from PDFs using pretrained models for layout analysis, OCR, and vision-language pairing. Additionally, we generate instruction data from extracted image-text pairs to enrich the training data.

We train Japanese LMMs with the PDF-derived data and evaluate them on the Japanese LMM Benchmark. Our results show that PDF-derived data significantly improves performance, with gains of 2.1\% to 13.8\% on Heron-Bench~\cite{inoue2024heron}. 
Furthermore, we conduct additional experiments to provide various insights into PDF-derived data's effectiveness. 
Our main contributions are as follows:
\begin{itemize}[noitemsep, topsep=0pt]
    \item We introduce a fully automated pipeline for extracting image-text pairs from PDFs using pretrained models, eliminating the need for manual annotation.
    \item We demonstrate that PDF-derived data can significantly improve Japanese LMM performance, achieving 2.1\% to 13.8\% gains on Heron-Bench.
    \item Through extensive experiments, we provide various insights into PDF-derived data's effectiveness. For example, we analyze the impact of PDF-derived data across different model sizes (3.8B, 8B, 14B) and evaluate the effectiveness of both image-text pairs and instruction data generated from images alone.
\end{itemize}

\section{Related Work}
\subsection{Extracting Image-text Pairs from PDFs}
Research on extracting images and their captions from PDFs, particularly scientific papers, has been actively explored~\cite{clark2015looking,clark2016pdffigures,siegel2018extracting,naiman2022figure,okamoto2023constructing}.
These studies typically perform the layout analysis~\cite{shen2021layoutparser} to locate image regions within a PDF, extract caption data from nearby text, and pair them together.
When pairing, some approaches use distance-based matching, considering that caption text is generally closer to the corresponding image than other text~\cite{okamoto2023constructing}.

However, to the best of our knowledge, no existing study has paired images with text other than the captions explicitly found in PDFs.
We aim to extract image-text pairs from PDFs without being limited to captions.
The closest existing work to our task is the identification of paragraphs that reference figures in scientific papers and summarizing their content to generate figure captions~\cite{huang2023summaries}.
However, this approach does not strictly pair images with non-caption text in PDFs, and its applicability is limited to scientific papers rather than general PDFs.
In contrast, our work extends beyond scientific papers to cover a broader range of general PDFs.

\subsection{Japanese LMM}\label{subsec:relatedwork-JA-LMM}
Recently, Japanese large multimodal models (LMMs) have been emerging based on English LMMs.
Proprietary LMMs have been improving their multilingual capabilities, achieving high performance in Japanese as well~\cite{openai2023gpt4,openai2024gpt4o,anthropic2024claude3,geminipro}.
Additionally, many open-source Japanese LMMs have been released~\cite{JapaneseInstructBLIPAlpha,JapaneseStableVLM,akiba2025evolutionary,BlipJapaneseStableLM,inoue2024heron,cyberagent2024llava,VILAjp}.

Most open-source LMMs follow the LLaVA~\cite{llava} approach, where a large language model (LLM) and a vision encoder are connected via a relatively shallow projector to form an LMM.
For training, some use in-house Japanese data~\cite{cyberagent2024llava}, while others rely on translated Japanese data~\cite{JapaneseInstructBLIPAlpha,JapaneseStableVLM,BlipJapaneseStableLM,inoue2024heron} and adopt a Japanese LLM as the base language model~\cite{cyberagent2024llava,JapaneseInstructBLIPAlpha,JapaneseStableVLM,BlipJapaneseStableLM,inoue2024heron}.
This approach enables the development of Japanese LMMs with decent performance.

Some models, such as Qwen-VL~\cite{bai2023qwen}, achieve strong performance in Japanese without using a Japanese LLM, instead leveraging a multilingual LLM.
VILA-jp~\cite{VILAjp} has further improved performance by utilizing interleaved data.
MangaLMM~\cite{baek2025mangalmm} specifically targets Japanese manga and demonstrates the potential of domain-specific Japanese LMMs.
However, no existing work has leveraged PDF data to enhance Japanese LMMs.
To achieve higher performance, we utilize PDF data in our approach.
Recently, several benchmarks have been introduced for evaluating Japanese LMMs, such as JDocQA~\cite{onami-etal-2024-jdocqa-japanese}, which focuses on Japanese document QA, and JMMMU~\cite{onohara2025jmmmu}, which considers Japanese culture.

\section{Harnessing PDF Data}\label{sec:PDF}
We aim to enhance the performance of Japanese LMMs using PDF data.
In this section, we describe the process of obtaining Japanese LMM training data from PDF data.

\subsection{Collecting PDF data}
The PDF dataset used in this study was collected from the Web based on URLs supplied by the Web ARchiving Project of the National Diet Library of Japan~\cite{NDL_WARP}.
The total number of PDFs exceeds 51.38 million.
However, we do not use all of these PDFs; instead, we select a subset through the following process.
The PDF data includes a wide variety of document types, not only academic or scientific papers but also newsletters, magazines, reports, posters, advertisements, campaign materials, pamphlets, brochures, manuals, and books.
Figure~\ref{fig:pdf-category} presents examples of various types of PDF data.

\begin{figure*}[t]
  \centering
  \includegraphics[width=0.93\linewidth]{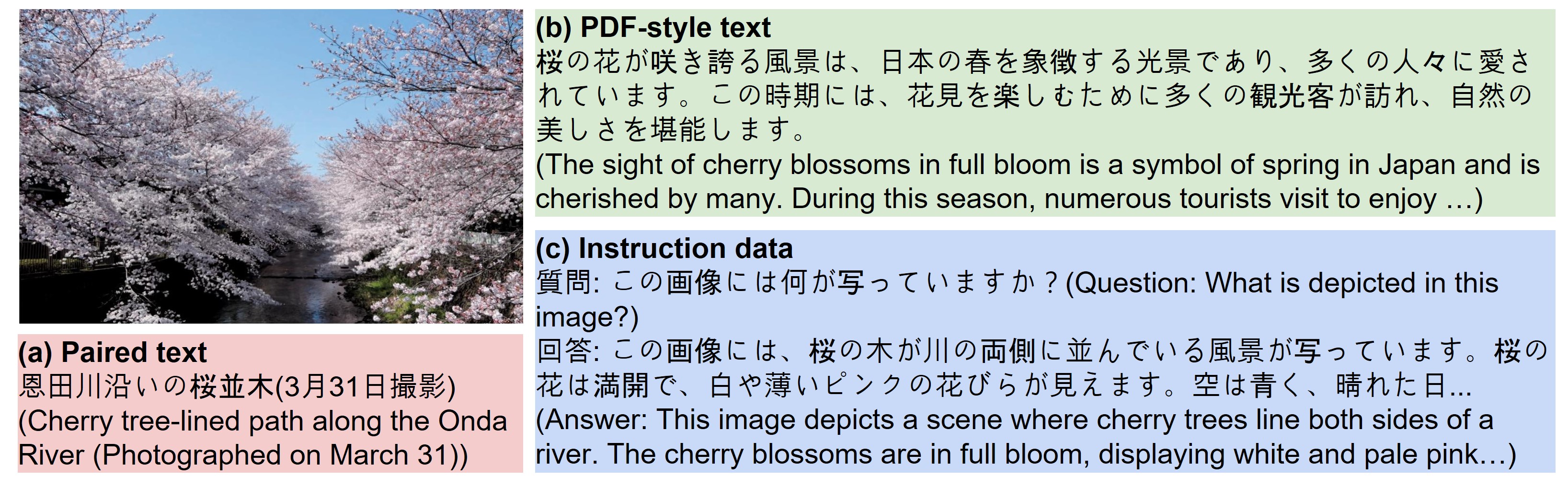}
  \vspace{-4mm}
  \caption{Example of an image with (a) paired text, (b) PDF-style text, and (c) instruction data}
  \label{fig:examples-text}
\end{figure*}

\subsection{Extracting Image-text Pairs}\label{subsec:extract-image-text}
To create training data for LMMs, we extract image-text pairs from PDF data.
The overall process is illustrated in Figure~\ref{fig:overview}.

\PAR{Selecting PDFs that Contain Images}
Before extracting images and text from PDFs, we first filter out PDFs that do not contain images.
A significant portion of the PDF data does not include any images, and many files contain only small logos or symbols rather than meaningful images.

After manually inspecting hundreds of PDFs, we observed that as the number of pages increases, PDFs tend to resemble books, where text dominates and images are scarce.
To address this, we select only PDFs with five or fewer pages.
Additionally, we found that images frequently appear on the first page of PDFs.
If an image is absent on the first page, subsequent pages are often image-free as well.
Thus, we extract only the first page from each selected PDF.

To detect whether a PDF contains images, we use a Python library.
There are several libraries available for this task~\cite{PyMuPDF,pdf2image,pdfminersix}.
Among them, we choose PyMuPDF~\cite{PyMuPDF}, which is widely used and offers both high speed and accuracy.
Using PyMuPDF, we identify PDFs that contain image data and filter out those without images.
As a result, we select 200K PDFs. Since we only use the first page of each PDF, this amounts to a total of 200K PDF pages.

\PAR{Extracting Image and Text through Layout Analysis and OCR}
The PyMuPDF library used for PDF selection directly reads a PDF and extracts images and text stored within the file.
However, this approach sometimes leads to issues.
For instance, PyMuPDF may extract images that are invisible to the human eye within a PDF.
Additionally, in some cases, it breaks down visible images and extracts them separately based on layout elements. 
For example, in some cases, the background and objects within an image are extracted separately.

To prevent such issues, we first convert a PDF into a JPEG image and then extract images and text from it.
This ensures that only images visible to the human eye are extracted.
To extract images and text, we use Surya~\cite{Surya}, a tool designed for PDF analysis based on image inputs.

Surya performs both layout analysis~\cite{shen2021layoutparser} and OCR.
First, layout analysis identifies image and text regions within the PDF.
Then, OCR is applied to the text regions to extract the text.
Through this process, we obtain both images and text.

Surya employs pretrained deep learning models for layout analysis and OCR, supporting over 90 languages, making it applicable to Japanese PDFs.
However, their performance is not perfect.
For example, despite processing Japanese text, it occasionally misidentifies characters as Hindi.
Additionally, we filter out images detected by Surya if their width or height is less than 50 pixels, as many non-image elements were mistakenly classified as images.
Further analysis of Surya’s pretrained model performance is provided in \S\ref{subsec:paired-text-train}.

\PAR{Pairing Image and Text}
For each image, we match the most semantically similar text.
Specifically, we embed the image using a vision encoder and OCR-extracted text using a text encoder.
We then compute the cosine similarity between the image and the text and select the text with the highest similarity as the paired text for the image.
For pairing, we use Japanese-Cloob~\cite{rinna-japanese-cloob-vit-b-16,youko-paper}, a pretrained vision-language model widely used in Japan (with 300K users last month), which follows a CLIP~\cite{clip}-like architecture.
Figure~\ref{fig:examples-text}(a) presents an example of the image-text pair.

\PAR{Filtering NSFW and PII data}
We found that PDF data sometimes contains (1) NSFW (Not Safe For Work) content and (2) PII (Personally Identifiable Information).
To filter out such data, we use GPT-4o-mini (\texttt{gpt-4o-mini-2024-07-18})~\cite{openai2024gpt4omini} to detect NSFW and PII content and exclude any data falling into these categories.
As a result, our dataset is largely free from such content and is expected to be safe for use.

\PAR{Generating PDF-style text} 
Pretrained models for layout analysis, OCR, and pairing are expected to perform well on scientific papers, as this domain has been extensively studied.
However, in our experiments, where we used a diverse range of PDFs beyond scientific papers, these pretrained models did not perform as effectively.
As a result, the quality of image-text pairs obtained using pretrained models was not always high.
Further details on this can be found in \S\ref{subsec:paired-text-train}.

This raises the question: ``What if we could extract image-text pairs from PDFs with higher accuracy?''
To explore this, we define PDF-style text and generate it using GPT-4o-mini, simulating an ideal paired text where each image is associated with a semantically relevant description.
Unlike direct image captions, PDF-style text does not explicitly describe the image in detail but instead provides a brief, indirect explanation derived from the surrounding text in the PDF.
For more details, refer to the prompt used for PDF-style text generation (Table~\ref{sup:prompt-pdfstyle}).

As shown in Figure~\ref{fig:examples-text}(b), the generated PDF-style text resembles the type of sentences commonly found in PDFs while providing an indirect description of the image.
The impact of training with PDF-style text is presented in Table~\ref{tab:paired}.

\subsection{Generating Instruction Data}\label{subsec:generate-inst-data}
Image-text pairs can be directly used for LMM training; however, their effectiveness was limited (see Table~\ref{tab:paired}).
Thus, instead of using them as they are, we followed the LLaVA~\cite{llava} approach and generated instruction data using GPT.

Strictly speaking, our method differs slightly from LLaVA.
At the time LLaVA was developed, GPT-4~\cite{openai2023gpt4} could not process image inputs, which likely explains why image data was not provided to GPT-4 during instruction generation.
However, by the time of our study, GPT-4o had been released, enabling image recognition.

Therefore, we directly feed images to GPT-4o-mini to generate instruction data. 
The paired text associated with each image is used as context information during instruction generation. 
For the prompt, we slightly modify the final part of the LLaVA prompt, ensuring that the responses are generated in Japanese.
The actual prompt used is provided in Table~\ref{sup:prompt-inst}.

Through our experiments, we found that when the paired text matched with an image is imperfect, it is more effective to generate instruction data using only the image, rather than including the paired text.
Further details on this can be found in \S\ref{subsec:paired-text-to-inst}.
Therefore, in our experiments, all instruction data, except for that used in Table~\ref{tab:pair-to-inst}, is generated using images only.
From 200K PDFs, we generate a total of 362K instruction tuning data.
Figure~\ref{fig:examples-text}(c) presents an example of the instruction data.

\section{Training Japanese LMM}
\begin{figure}[t]
  \centering
  \includegraphics[width=\columnwidth]{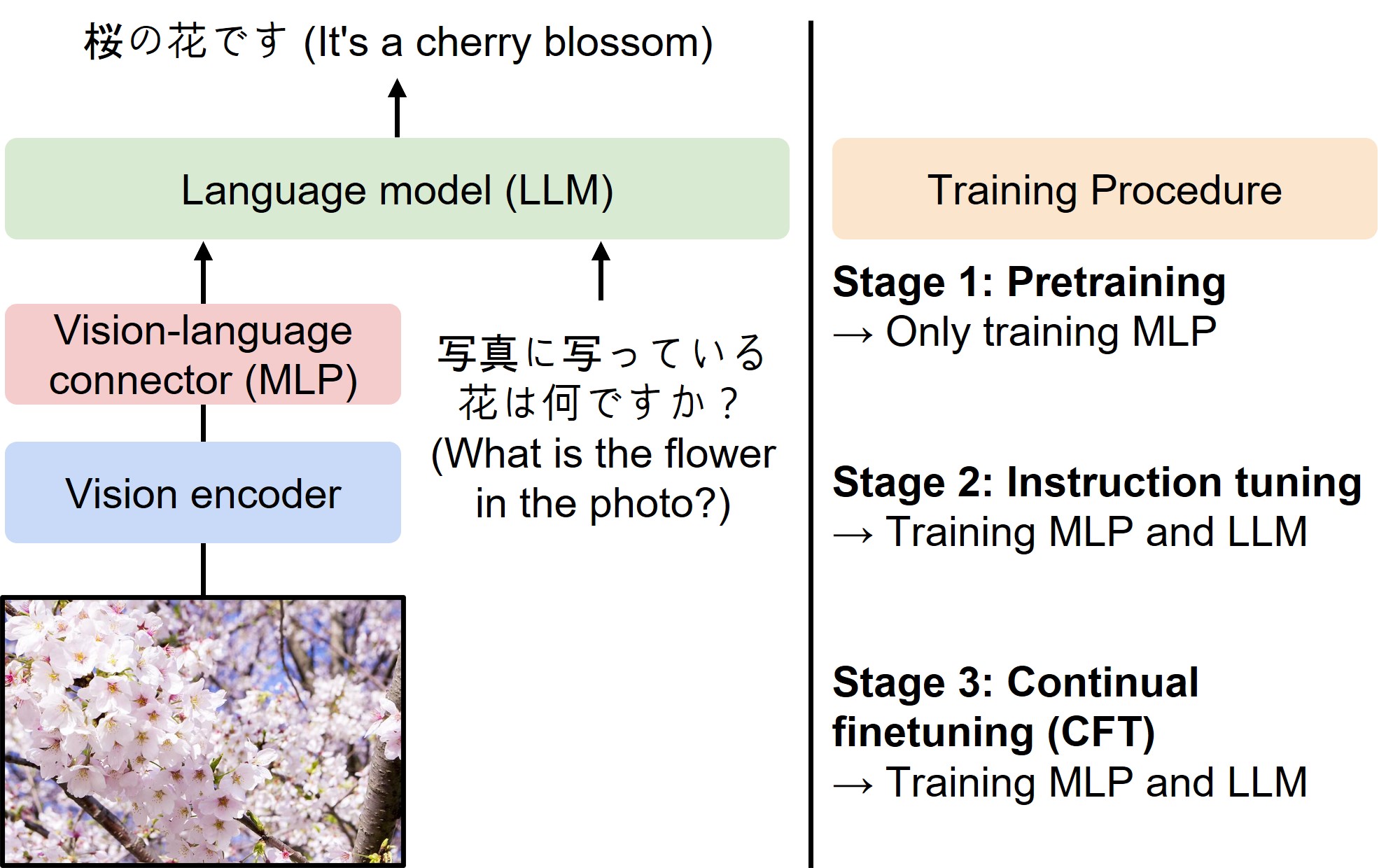}
  \vspace{-6mm}
  \caption{\textbf{Training LMM.} 
  We use the LLaVA1.5~\cite{llava1.5} framework in our experiments. 
  The training consists of 3 stages: pretraining, instruction tuning, and continual finetuning (CFT).}
  \label{fig:llava}
\end{figure}

Recently, various open-source Japanese LMMs have been released~\cite{JapaneseInstructBLIPAlpha,JapaneseStableVLM,akiba2025evolutionary,BlipJapaneseStableLM,inoue2024heron,cyberagent2024llava,VILAjp}.
Among them, we adopt the widely used LLaVA~\cite{llava} framework to evaluate the effectiveness of PDF data, specifically using LLaVA1.5~\cite{llava1.5}.
Figure~\ref{fig:llava} illustrates the LLaVA1.5 framework.
Most hyperparameters follow the original LLaVA1.5 settings, with a few modifications.
We replace the vision encoder CLIP (clip-vit-large-patch14-336)~\cite{clip} with SigLIP (siglip-so400m-patch14-384)~\cite{SigLIP} and experiment with different large language models (LLMs) instead of Vicuna-7B~\cite{vicuna2023}.
Details on the LLM selection will be discussed later.
For training, we employ LoRA~\cite{lora} for parameter-efficient finetuning.

\begin{table}[t] 
  \tabcolsep=0.15cm
    \begin{center}
    \vspace{-1mm}
        \begin{adjustbox}{width=0.83\linewidth}
        \begin{tabular}{@{}l|crr@{}}
            \toprule
            \textbf{Dataset} & \textbf{Stage} &  \textbf{Count} \\
            \midrule
            LLaVA-Pretrain-JA & 1 & 558K \\
            LLaVA-v1.5-Instruct-620K-JA & 2 & 620K \\
            Instruct-from-200K PDF & 3 & 362K \\ 
            \bottomrule
        \end{tabular}
        \end{adjustbox}
    \vspace{-2mm}
    \caption{Details of the Japanese LMM training sets.}
    \label{tab:traindata}
    \end{center}
\end{table} 

\subsection{Training Procedure}
LLaVA1.5 training consists of two stages, aiming to integrate the pretrained vision encoder and pretrained LLM to create an LMM that effectively handles vision input.

\PAR{Stage 1: Pretraining}
In this stage, only the vision-language connector (MLP) is trained using image-text pairs, linking the vision encoder with the LLM.

\PAR{Stage 2: Instruction Tuning}
Using visual instruction data, both the MLP and LLM are instruction-tuned to improve multimodal understanding.

\PAR{Stage 3: Continual Fine-Tuning (CFT)} 
In our study, we introduce an additional CFT stage after Stages 1 and 2.
Here, we perform CFT on both the MLP and LLM using PDF data.

\begin{table}[t] 
  \tabcolsep=0.15cm
    \begin{center}
    \vspace{-1mm}
        \begin{adjustbox}{width=0.83\linewidth}
        \begin{tabular}{@{}l|cc@{}}
            \toprule
            \textbf{Dataset} & \textbf{\#Questions} & \textbf{\#Images}  \\
            \midrule
            JA-LLaVA-Bench (COCO) & 90 & 30 \\
            JA-LLaVA-Bench (Wild) & 60 & 24 \\ 
            Heron-Bench & 103 & 21 \\
            \bottomrule
        \end{tabular}
        \end{adjustbox}
    \vspace{-2mm}
    \caption{Details of Japanese LMM evaluation sets.}
    \label{tab:evaldata}
    \end{center}
\end{table}

\begin{table*}[t] 
    \tabcolsep=0.13cm  
    \centering
    \begin{adjustbox}{width=\linewidth}
     \begin{tabular}[t]{@{}ll|cccc|cccc|cccc@{}}
        \toprule
        & & \multicolumn{4}{c|}{\textbf{JA-LLaVA-Bench (COCO)}} & \multicolumn{4}{c|}{\textbf{JA-LLaVA-Bench (Wild)}} & \multicolumn{4}{c}{\textbf{Heron-Bench}}\\ 
        & \textbf{Method}  & Detail & Conv & Complex & Avg. & Detail & Conv & Complex & Avg. & Detail & Conv & Complex & Avg. \\
        \midrule
        \parbox[t]{3mm}{\multirow{10}{*}{\rotatebox[origin=c]{90}{Results from Heron-Bench}}}
         & GPT-4V & 88.0 & 84.7 & 97.5 & 90.1 & 89.9 & 93.1 & 99.1 & 94.1 & 83.3 & 77.5 & 78.3 & 79.7 \\
 & Claude 3 Opus & 67.9 & 70.4 & 95.3 & 77.9 & 87.4 & 73.4 & 94.6 & 85.1 & 74.5 & 68.4 & 77.7 & 73.6 \\
 & Gemini Pro & 65.3 & 83.0 & 75.4 & 74.6 & 61.7 & 84.2 & 84.2 & 76.7 & 55.6 & 64.3 & 64.0 & 61.3 \\
\cline{2-14}
 & LLaVA 1.6 7B & 60.3 & 83.7 & 60.6 & 68.2 & 36.7 & 44.7 & 53.4 & 44.9 & 30.9 & 37.3 & 31.0 & 33.1 \\
 & LLaVA 1.5 7B & 71.1 & 83.7 & 69.6 & 74.8 & 49.2 & 48.7 & 54.7 & 50.8 & 42.4 & 45.9 & 35.5 & 41.3 \\
 & Qwen-VL 7B & 78.0 & 81.0 & 82.2 & 80.4 & 55.9 & 49.7 & 56.4 & 54.0 & 46.3 & 50.6 & 52.3 & 49.7 \\
 & Japanese StableVLM 7B & 18.9 & 54.8 & 24.1 & 32.6 & 26.0 & 24.8 & 29.2 & 26.7 & 25.2 & 51.2 & 37.8 & 38.1 \\
 & EvoVLM-JP-v1 7B & 61.0 & 75.7 & 71.0 & 69.2 & 49.6 & 65.5 & 54.2 & 56.4 & 50.3 & 44.4 & 40.5 & 45.1 \\
 & Heron BLIP v1 7B & \textbf{84.8} & 94.3 & \textbf{89.5} & \textbf{89.5} & 45.5 & 32.9 & 56.9 & 45.1 & 49.1 & 41.5 & 45.7 & 45.4 \\
 & Heron GIT 7B & 83.0 & 78.2 & 91.1 & 84.1 & 41.0 & 39.9 & 54.6 & 45.2 & 42.8 & 54.2 & 43.5 & 46.8 \\
        \midrule
        \parbox[t]{3mm}{\multirow{4}{*}{\rotatebox[origin=c]{90}{Our results}}}
        & PDF-JLMM 8B & 81.6 & \textbf{95.9} & 87.1 & 88.2 & 69.3 & 50.6 & \textbf{77.6} & 65.8 & \textbf{70.1} & \textbf{62.3} & \textbf{65.0} & \textbf{65.8} \\ 
        & LLaVA1.5-Llama3 8B & 82.8 & 90.6 & 87.2 & 86.9 & 60.5 & 38.7 & 71.5 & 56.9 & 68.7 & 60.1 & 56.1 & 61.6 \\ 
        & LLaVA1.5-Phi3-mini 3.8B & 77.3 & 89.8 & 83.7 & 83.6 & 66.7 & 36.5 & 66.8 & 56.7 & 61.6 & 61.2 & 48.5 & 57.1 \\ 
        & LLaVA1.5-Phi3-medium 14B & 84.7 & 89.6 & 86.2 & 86.8 & \textbf{77.0} & \textbf{68.2} & 77.0 & \textbf{74.1} & 62.0 & 56.1 & 54.1 & 57.4 \\ 
        \bottomrule
    \end{tabular}
    \end{adjustbox}
    \vspace{-2mm}
    \caption{\textbf{Main results.}
    Our models trained with PDF data achieve high performance across all benchmarks except JA-LLaVA-Bench (COCO).
    Our LLaVA1.5-based models are named after their backbone LLMs (e.g., LLaVA1.5-Llama3).
    PDF-JLMM refers to LLaVA1.5 with Swallow as the backbone LLM.
The best-performing open-source models are highlighted in \textbf{bold}.
    }
    \label{tab:main_results}
\end{table*}

\PAR{Training Data}
Table~\ref{tab:traindata} presents the details of the training data.
In Stage 1 (Pretraining) and Stage 2 (Instruction Tuning), we use the Japanese-translated version of the original LLaVA training data~\cite{inoue2024heron}, translated using DeepL~\cite{DeepL}.
Specifically, Stage 1 uses 558K samples from LLaVA-Pretrain-JA~\cite{llava_pretrain_ja}, Stage 2 uses 620K samples from LLaVA-v1.5-Instruct-620K-JA~\cite{llava_instruct_620k_ja}, and Stage 3 uses 362K PDF-derived samples created in \S\ref{sec:PDF}.

\PAR{Elapsed Time for Training}
LLaVA1.5 is trained for one epoch per stage.
Training LLaVA1.5 with a Llama3 8B-based LLM using four NVIDIA A100 GPUs took about 11 hours for Stage 1, 42 hours for Stage 2, and 19 hours for Stage 3.

\subsection{LLM Selection}\label{subsec:llm}
\PAR{Japanese LLM}
We train LLaVA1.5 using three well-known Llama3-8B-based Japanese LLMs:
Suzume (suzume-llama-3-8B-japanese~\cite{suzume-paper}), ELYZA (Llama-3-ELYZA-JP-8B~\cite{elyzallama2024}), and Swallow (Llama-3-Swallow-8B-Instruct-v0.1~\cite{swallow-paper})
We build our main model by finetuning LLaVA1.5, using Swallow as the base Japanese LLM, with Japanese PDF data.
We name this model \textbf{PDF-JLMM} (PDF-based Japanese Large Multimodal Model).

\PAR{General (Non-Japanese) LLM}
To verify whether Japanese PDF data is effective in adapting a general (non-Japanese) LLM into a Japanese LMM, we use three non-Japanese LLMs:
Llama3 (Llama-3-8B-instruct~\cite{llama3.1}), Phi3-mini (Phi-3-mini-4k-instruct~\cite{abdin2024phi}, 3.8B parameters), and Phi3-medium (Phi-3-medium-4k-instruct~\cite{abdin2024phi}, 14B parameters).

\section{Experiments and Analysis}

\subsection{Evaluation Metric} 
To evaluate Japanese LMMs, we adopt the evaluation method used in Heron-Bench~\cite{inoue2024heron}, a standard Japanese LMM benchmark.
The authors of Heron-Bench provide three evaluation datasets: JA-LLaVA-Bench (COCO) and JA-LLaVA-Bench (Wild), which are Japanese translations of LLaVA-Bench\cite{llava}, and Heron-Bench, specifically designed for Japanese evaluation.
The details of them are presented in Table~\ref{tab:evaldata}.

Heron-Bench follows the same score calculation method as LLaVA-Bench.
First, GPT-4 (\texttt{gpt-4-0125-preview})~\cite{openai2023gpt4} generates reference answers using the question's context.
Then, GPT-4 evaluates both LMM's answers and the reference answers using the LLM-as-a-judge approach~\cite{judge}.
The final score is calculated as the ratio (\%) of the average score of the LMM’s answers to the average score of GPT-4’s reference answers.
A score of 100\% indicates performance on par with GPT-4, while scores above 100\% suggest that the LMM outperforms GPT-4.

\PAR{LMMs used for comparison}
We use three proprietary LMMs—GPT-4V~\cite{openai2023gpt4}, Claude 3 Opus~\cite{anthropic2024claude3}, and Gemini Pro~\cite{geminipro}—along with seven open-source LMMs—LLaVA 1.6 7B~\cite{liu2024llavanext}, LLaVA 1.5 7B~\cite{llava}, Qwen-VL 7B~\cite{bai2023qwen}, Japanese StableVLM 7B~\cite{JapaneseStableVLM}, EvoVLM-JP-v1-7B~\cite{akiba2025evolutionary}, Heron BLIP v1 (620k)~\cite{BlipJapaneseStableLM}, and Heron GIT~\cite{inoue2024heron}.

\subsection{Main Result}
Table~\ref{tab:main_results} presents the results of our model trained on PDF-derived data.
Compared to existing models from the Heron-Bench paper, our PDF-JLMM outperforms most open-source Japanese LMMs.
It lags behind Heron BLIP v1 7B by only 1.3\% on JA-LLaVA-Bench (COCO).
On JA-LLaVA-Bench (Wild), it achieves a performance of 65.8\%, surpassing the previous best of 56.4\% by 9.4\%.
For Heron-Bench, it outperforms the previous best of 49.7\% by a significant margin of 16.1\%.
Additionally, our LLaVA1.5-Llama3, LLaVA1.5-Phi3-mini, and LLaVA1.5-Phi3-medium also achieve higher performance than existing models.
These results demonstrate that \textit{utilizing PDF data effectively enhances model performance}.

In the following subsections, we conduct additional experiments on training with PDF-derived data and provide a component-wise analysis.

\subsection{Is Using PDF-derived Data Effective?}
\begin{table}[t]
    \centering
    \begin{adjustbox}{width=\linewidth}
    \begin{tabular}{@{}clccc@{}}
        \toprule
        \textbf{LLM} & \textbf{Stage} & \textbf{L-COCO} & \textbf{L-Wild} &  \textbf{Heron} \\
        \midrule
        \parbox[t]{0mm}{\multirow{6}{*}{\rotatebox[origin=c]{90}{Swallow 8B}}}
        & 1. Pretraining & 30.6 & 17.4 & 22.6 \\ 
        & 2. Instruction tuning & 84.0 & 59.8 & 54.7 \\ 
        & 3. CFT on 50K PDF & 87.3 & 61.6 & 65.7 \\ 
        & \textbf{3. CFT on 100K PDF} & \textbf{88.2} & \textbf{65.8} & \textbf{65.8} \\ 
        & 3. CFT on 150K PDF & 88.1 & 65.5 & 63.8 \\ 
        & 3. CFT on 200K PDF & 86.6 & 64.7 & 64.6 \\ 
        \midrule
        \midrule
        \parbox[t]{0mm}{\multirow{6}{*}{\rotatebox[origin=c]{90}{Llama3 8B}}}
        & 1. Pretraining & 26.8 & 16.5 & 23.2 \\ 
        & 2. Instruction tuning & 84.4 & 57.0 & 54.8 \\ 
        & 3. CFT on 50K PDF & 86.5 & 54.0 & 58.7 \\ 
        & 3. CFT on 100K PDF & 86.3 & \textbf{57.1} & 61.0 \\ 
        & 3. CFT on 150K PDF & 85.1 & 53.9 & \textbf{61.8} \\ 
        & \textbf{3. CFT on 200K PDF} & \textbf{86.9} & 56.9 & 61.6 \\ 
        \midrule
        \midrule
        \parbox[t]{0mm}{\multirow{6}{*}{\rotatebox[origin=c]{90}{Phi3-mini 3.8B}}}
        & 1. Pretraining & 21.8 & 13.5 & 18.4 \\ 
        & 2. Instruction tuning & 82.7 & 50.6 & 43.3 \\ 
        & 3. CFT on 50K PDF & 83.0 & 52.3 & 51.9 \\ 
        & 3. CFT on 100K PDF & 83.3 & 52.0 & 53.0 \\ 
        & \textbf{3. CFT on 150K PDF} & 83.6 & \textbf{56.7} & \textbf{57.1} \\ 
        & 3. CFT on 200K PDF & \textbf{83.8} & 48.2 & 54.3 \\ 
        \midrule
        \midrule
        \parbox[t]{0mm}{\multirow{6}{*}{\rotatebox[origin=c]{90}{Phi3-medium 14B}}}
        & 1. Pretraining & 27.5 & 16.0 & 23.4 \\ 
        & 2. Instruction tuning & 86.3 & 71.4 & 54.2 \\ 
        & 3. CFT on 50K PDF & 87.4 & 65.2 & \textbf{58.8} \\ 
        & 3. CFT on 100K PDF & 85.9 & 66.8 & 56.3 \\ 
        & \textbf{3. CFT on 150K PDF} & 86.8 & \textbf{74.1} & 57.4 \\ 
        & 3. CFT on 200K PDF & \textbf{88.5} & 70.7 & 58.1 \\ 
        \bottomrule
    \end{tabular}
    \end{adjustbox}
    \vspace{-2mm}
    \caption{\textbf{Results for each stage and increasing amounts of PDF.}
    CFT on PDF-derived data is effective.
L-COCO, L-Wild, and Heron correspond to JA-LLaVA-Bench (COCO), JA-LLaVA-Bench (Wild), and Heron-Bench, respectively.
The average value per benchmark is shown.
For each LLM, the highest-performing stage and value are highlighted in \textbf{bold}.
    }
    \label{tab:stage-per-LLM}
\end{table}

Table~\ref{tab:stage-per-LLM} presents the results of LLaVA1.5 training for each LLM across different stages. 
Overall, continual fine-tuning (CFT) on PDF-derived data is effective.
Immediately after Stage 1 (Pretraining), all LLMs exhibit relatively low performance.
After Stage 2 (Instruction tuning), Heron-Bench performance improves by approximately 20\%–30\%.
After Stage 3 (CFT on PDF data), Heron-Bench performance further increases by at least 2.1\% (Phi3-medium) and up to 13.8\% (Phi3-mini).

\begin{figure}[t]
  \includegraphics[width=\columnwidth]{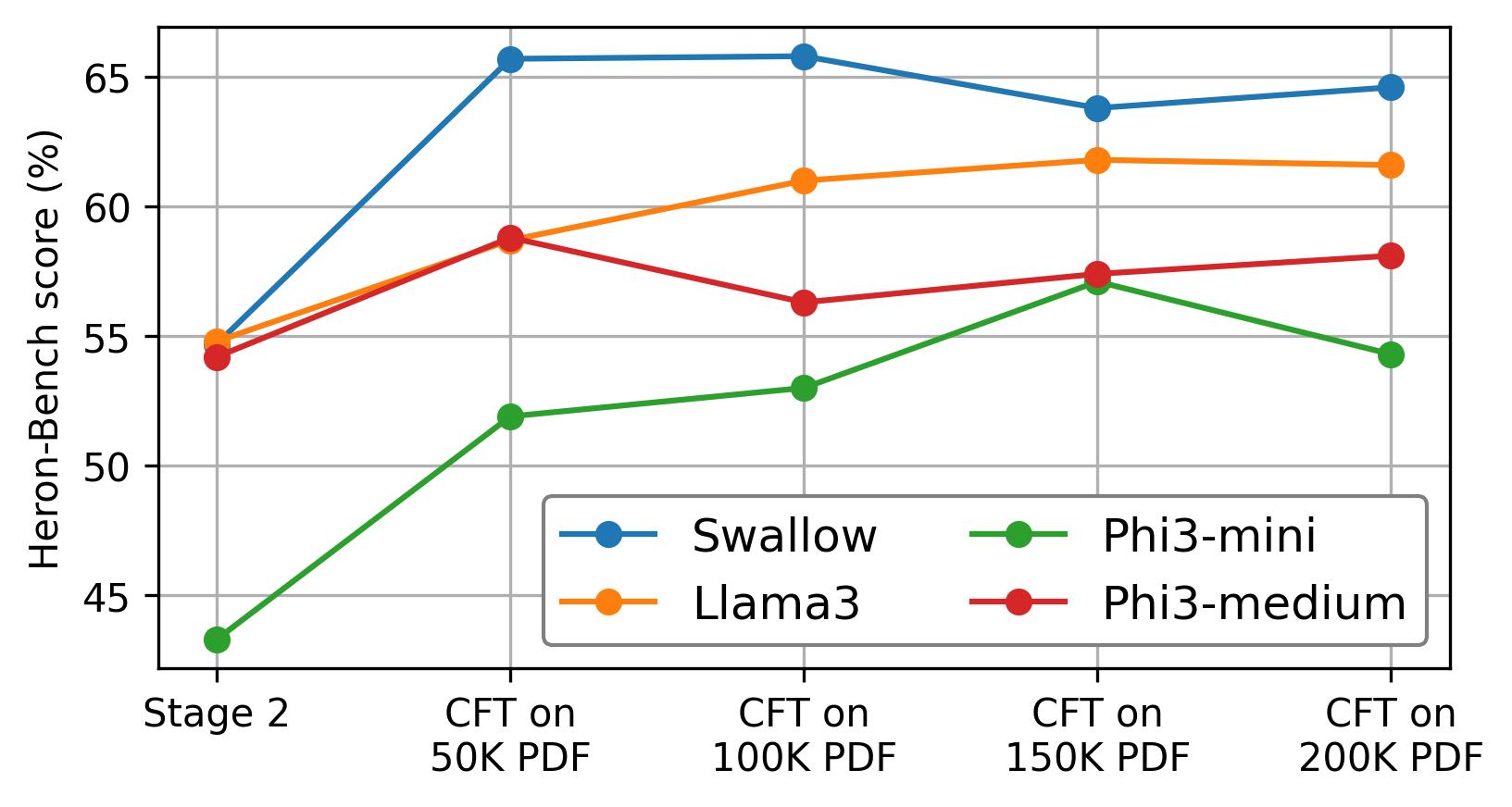}
  \vspace{-6mm}
  \caption{\textbf{Performance on Heron-Bench with CFT on PDF.}
CFT improves performance.
However, more data does not always lead to continuous performance gains.
  }
  \label{fig:heron-score}
\end{figure}

For Stage 3, we also show experiments with increasing amounts of PDF data, where every additional 50K PDFs adds approximately 90K new instruction data.
Figure~\ref{fig:heron-score} demonstrates the performance improvement on Heron-Bench when applying CFT on PDF data, highlighting the effectiveness of PDF data.
However, performance does not always increase as the amount of PDF data grows.
Performance does not improve in direct proportion to the amount of data, indicating that scaling is not strictly linear.

Additionally, Japanese PDF data is also effective in adapting a general (non-Japanese) LLM into a Japanese LMM.
To verify this, we conducted the same experiment using Llama3 and Phi3, both non-Japanese LLMs, as described in \S\ref{subsec:llm}.
As shown in Table~\ref{tab:main_results} and Figure~\ref{fig:heron-score}, PDF data is effective for training these models as Japanese LMMs.

Moreover, PDF data is effective across various model sizes.
It benefits not only models based on Llama3 8B but also Phi3-mini (3.8B) and Phi3-medium (14B).
However, in the case of Phi3, even at a larger model size of 14B, its performance lags behind the 8B Llama3-based model.
This may be due to its limited Japanese vocabulary. Phi3 contains only 837 Japanese vocabulary tokens.
Considering that commonly used kanji in Japanese amount to around 2,000~\cite{joyo_kanji_wiki}, 837 vocabulary tokens are likely insufficient for effectively handling Japanese.
As a result, despite having a larger model size of 14B, its performance remains comparable to that of the 8B Llama3 model.

\subsection{Which Japanese LLM Performs Best?}
\begin{table}[t]
    \centering
    \begin{adjustbox}{width=\linewidth}
    \begin{tabular}{@{}lccc@{}}
        \toprule
        \textbf{Method} & \textbf{L-COCO} & \textbf{L-Wild} &  \textbf{Heron} \\
        \midrule
        LLaVA1.5-Suzume 8B & 83.1 & 55.4 & 51.9 \\ 
        LLaVA1.5-ELYZA 8B & 81.8 & 55.1 & 53.5 \\ 
        LLaVA1.5-Swallow 8B & \textbf{84.0} & \textbf{59.8} & \textbf{54.7} \\ 
        \bottomrule
    \end{tabular}
    \end{adjustbox}
    \vspace{-2mm}
    \caption{\textbf{LLaVA1.5 results for each Japanese LLM.}
The results are from training up to Stage 1 and 2.
    }
    \label{tab:ja-llm}
\end{table}

We train LLaVA1.5 up to Stage 1 and 2 using the three commonly used Japanese LLMs introduced in \S\ref{subsec:llm} (Suzume, ELYZA, and Swallow).
Table~\ref{tab:ja-llm} presents the results.
We find that Swallow achieves the best performance across all three benchmarks, and therefore, we select Swallow as the Japanese LLM for this study.

\subsection{What If We Use Raw Image-Text Pairs Without Generating Instruction Data?}\label{subsec:paired-text-train}
\begin{table}[t]
    \centering
    \begin{adjustbox}{width=0.8\linewidth}
    \begin{tabular}{@{}lccc@{}}
        \toprule
        \textbf{Method} & \textbf{L-COCO} & \textbf{L-Wild} &  \textbf{Heron} \\
        \midrule
        Stages 1 and 2 & \textbf{84.0} & \textbf{59.8} & 54.7 \\ 
        \midrule
        Top 1 & 77.0 & 37.4 & 40.0 \\ 
        Top 3 & 72.3 & 31.3 & 34.7 \\ 
        Top 5 & 62.6 & 26.6 & 22.8 \\ 
        Neighbor & 81.7 & 39.9 & 46.4 \\ 
        \midrule
        PDF-style text & 81.5 & 56.5 & \textbf{65.5} \\
        \bottomrule
    \end{tabular}
    \end{adjustbox}
    \vspace{-2mm}
    \caption{\textbf{Results from training with raw image-text pairs.}
When performing CFT using only image-text pairs, the overall performance is lower than the baseline.
    }
    \label{tab:paired}
\end{table}

Instead of generating instruction data from extracted image-text pairs, an LMM can be trained directly on them. We train PDF-JLMM with 50K PDF data using extracted image-text pairs. Table~\ref{tab:paired} presents the results.

The results show that performance generally decreases compared to LLaVA1.5 trained only up to Stages 1 and 2.
In the table, Top 1, 3, 5 refer to results obtained using the top 1, 3, or 5 texts ranked by cosine similarity when pairing with an image.
Neighbor denotes the setting where, in addition to the Top 1 text, one preceding and one following text from the same PDF are also included.
Even when using Top 3, Top 5, or Neighbor, performance is still lower than that of LLaVA1.5 trained only up to Stage 2.
This suggests that image-text pairs extracted solely using pretrained models are not effective as training data.

There are two major possible reasons for this performance drop:
(1) Many PDFs inherently contain little text that directly describes the images.
(2) The limitations of pretrained models. The quality of image-text pairs may have been low due to insufficient performance in layout analysis, OCR, and pairing tasks.
Since there is no annotation data for these PDFs, it is difficult to rigorously determine which factor is the primary cause.
However, upon manually inspecting hundreds of examples, we observed that the OCR-extracted text was often inaccurate.
Many extracted texts contained unintended line breaks, leading to broken words or sentences, and complex kanji characters were frequently misrecognized.

These results led us to question: ``What if image-text pairs were accurately extracted from PDFs?''
To explore this, we generated an ideal paired text for each image using GPT-4o-mini.
This approach, referred to as PDF-style text, is described in \S\ref{subsec:extract-image-text}.

Using PDF-style text significantly outperforms raw image-text pairs.
Compared to Stages 1 and 2, it achieves a 10.8\% improvement on Heron-Bench.
These findings suggest that if text data is properly extracted from PDFs, training solely on image-text pairs could lead to further performance gains.

\subsection{Is Paired Text Effective for Generating Instruction Data?}\label{subsec:paired-text-to-inst}
When generating instruction data, we use paired text matched with images as context.
However, the effectiveness of this paired text remains uncertain.
Since the text extracted from PDFs is often imperfect and contains noise, its usefulness for training may be uncertain.
To investigate this, we generate instruction data using different data sources derived from 50K PDFs and present the results in Table~\ref{tab:pair-to-inst}.
We compare three cases: using only images, images with paired text, and images with PDF-style text.

The key takeaway from our experiment is that instruction data generated using only images performed best.
This suggests that even extracting only image data from PDFs is valuable, and when paired text is inaccurate, using images alone can yield better performance.
Thus, for experiments using 200K PDFs, we generated instruction data using only images.

From these results, we conclude that paired text containing noise is not effective.
However, the results also show that PDF-style text improves performance compared to paired text.
This implies that if text data is more accurately extracted from PDFs, the quality of instruction data generated from image-text pairs can also improve.

\begin{table}[t]
    \centering
    \begin{adjustbox}{width=\linewidth}
    \begin{tabular}{@{}lccc@{}}
        \toprule
        \textbf{Data source for instruction} & \textbf{L-COCO} & \textbf{L-Wild} &  \textbf{Heron} \\
        \midrule
        Image & 87.3 & 61.6 & \textbf{65.7} \\ 
        Image and paired text & \textbf{87.5} & 60.0 & 63.9 \\ 
        Image and PDF-style text & 87.2 & \textbf{63.0} & 64.0 \\
        \bottomrule
    \end{tabular}
    \end{adjustbox}
    \vspace{-2mm}
    \caption{\textbf{Comparison of data sources for generating instruction data.} 
    PDF-JLMM is used in experiments.
    }
    \label{tab:pair-to-inst}
\end{table}

\subsection{Comparison with the Translated Datasets}
We compare the PDF-derived data with Japanese-translated versions of existing English instruction data. 
The 620K instruction data (LLaVA-v1.5-Instruct-620K-JA) used in Stage 2 is considered a different Japanese multimodal dataset, constructed by translating English instruction data.
We compare it with our Instruct-from-200K PDF dataset (362K instruction data).
Instead of using LLaVA-v1.5-Instruct-620K-JA in Stage 2, we train the model with our Instruct-from-200K PDF dataset. 
Table~\ref{tab:llava-data-vs-pdf} shows the results.
LLaVA1.5-Swallow 8B trained with the PDF-derived data outperforms the one trained with LLaVA-v1.5-Instruct-620K-JA, even with fewer samples. 
This result suggests that the performance gains are not merely due to large-scale instruction tuning, but rather from leveraging instruction data constructed from Japanese PDFs containing culturally grounded content.

\begin{table}[t]
    \centering
    \begin{adjustbox}{width=\linewidth}
    \begin{tabular}{@{}lccc@{}}
        \toprule
        \textbf{Data source for Stage 2} & \textbf{L-COCO} & \textbf{L-Wild} &  \textbf{Heron} \\
        \midrule
        LLaVA-v1.5-Instruct-620K-JA	 & 84.0 & 59.8 & 54.7 \\ 
        Instruct-from-200K PDF & \textbf{88.1} & \textbf{72.7} & \textbf{70.0} \\ 
        \bottomrule
    \end{tabular}
    \end{adjustbox}
    \vspace{-2mm}
    \caption{\textbf{Comparison with the translated data, LLaVA-v1.5-Instruct-620K-JA.}
    }
    \label{tab:llava-data-vs-pdf}
\end{table}

We also compare the PDF-derived data with other datasets that we translated into Japanese.  
Specifically, we translated Vision-Flan~\cite{xu2024visionflan} (186K instruction data) and Image-Textualization~\cite{pi2024imagetextualization} (99.6K instruction data)—subsets of LLaVA OneVision~\cite{li2024llava}—into Japanese using GPT-4o-mini.
For translation, we use the prompt in Table~\ref{sup:prompt-translate}.
We use these datasets for CFT in Stage 3.
Table~\ref{tab:translated-llava-ov} shows the results.  
The results show that simply translating effective English data into Japanese may even lead to performance degradation on Japanese benchmarks.  
This underscores the importance of using Japanese PDF data that contains authentic, culturally grounded content.

Our approach differs from translated English datasets in that we extract images containing authentic Japanese content directly from Japanese PDFs. 
For example, as shown in Figure~\ref{fig:heronqa1} of the qualitative analysis in the supplementary materials, the model trained only up to Stage 2 (before incorporating our PDF-derived data) describes ``cherry blossoms''—a culturally symbolic flower in Japan—as ``white flowers'', whereas the model further trained on PDF-derived data correctly refers to it as ``cherry blossoms''. 
We believe this illustrates a key distinction from translated data, as our method helps the model learn cultural concepts grounded in native Japanese contexts rather than relying on translations.

\begin{table}[t]
    \centering
    \begin{adjustbox}{width=\linewidth}
    \begin{tabular}{@{}lccc@{}}
        \toprule
        \textbf{Data source for Stage 3} & \textbf{L-COCO} & \textbf{L-Wild} &  \textbf{Heron} \\
        \midrule
        100K PDF (181K inst.) & \textbf{88.2} & \textbf{65.8} & \textbf{65.8} \\ 
        Vision-Flan (VF) & 37.8 & 29.0 & 40.2 \\ 
        Image-Textualization (IT) & 81.6 & 43.3 & 46.3 \\
        VF + IT & 56.7 & 35.5 & 39.9 \\
        \bottomrule
    \end{tabular}
    \end{adjustbox}
    \vspace{-2mm}
    \caption{\textbf{Comparison with the translated subset of LLaVA-OneVision.}
    LLaVA1.5-Swallow is used in experiments.
    }
    \label{tab:translated-llava-ov}
\end{table}

\section{Conclusion}
We explore the use of Japanese PDF data to enhance LMM training and develop a fully automated pipeline for extracting image-text pairs. Our experiments show significant performance gains by incorporating PDF-derived data, with up to 13.8\% improvement on Heron-Bench.
Further analysis confirms the effectiveness of PDF-derived data across different model sizes and its potential to complement existing multimodal datasets. 
These findings provide valuable insights into leveraging PDF data for LMM training and highlight its promise as a multimodal resource. 
While our focus is on Japanese, we believe this approach is applicable to other languages as well, and we hope it fosters further research toward improving LMMs across diverse languages.

\section*{Limitations}
Following LLaVA’s approach of generating instruction data using GPT, we used GPT-4o-mini to generate instruction data from PDF data.
While this approach is effective, it is dependent on GPT.
To generate LMM training data without relying on GPT, high-quality image-text pair data is essential.
Achieving this requires improving the performance of text extraction models for PDFs.

Currently, scaling beyond 100K PDFs has been challenging.
For future work, we plan to investigate the underlying causes—whether the bottleneck lies in data quality, model capacity, suboptimal training settings (e.g., learning rate, number of epochs), or the limitations of existing test data.
By addressing these factors, we believe that scaling to larger PDF datasets will become more feasible.

\section*{Acknowledgments}
We acknowledge the support of the Research and Development Center for Large Language Models at the National Institute of Informatics for providing the PDF dataset. In particular, we would like to thank Professor Daisuke Kawahara for his guidance regarding copyright and permissions.
This work was supported by JSPS KAKENHI Grant Number 24K23882 and by the NVIDIA Academic Grant Program.

\bibliography{custom}

\newcommand\beginsupplement{%
        \setcounter{table}{0}
        \renewcommand{\thetable}{\Alph{table}}%
        \setcounter{figure}{0}
        \renewcommand{\thefigure}{\Alph{figure}}%
     }
\beginsupplement
\appendix
\section{Prompts Used in Our Experiments}\label{supsec:prompt}
This section provides the prompts used in our experiments.
Table~\ref{sup:prompt-pdfstyle}, Table~\ref{sup:prompt-inst}, and Table~\ref{sup:prompt-translate} show the prompts used to generate PDF-style text, instruction data, and Japanese translations of English datasets, respectively.

\begin{table*}[t]
\begin{tcolorbox}[colback=gray!10, colframe=black, rounded corners]
You are an AI visual assistant, and you are seeing a single image. Generate a passage that resembles text commonly found in PDF documents and is relevant to the given image. The provided image is extracted from a PDF, but no additional context, such as the document’s text or structure, is available.\\
PDF-style text generally has the following characteristics:\\
1. No explicit captions or minimal captions: Instead of directly describing the image, related text may naturally integrate into the document’s content.\\
2. Indirect descriptions: The text does not explicitly reference the image but provides supporting information that the image complements.
\\\\
To keep the text concise, generate only 1 to 2 sentences per image, ensuring it aligns with common PDF writing styles. \\
You must respond in Japanese.
\end{tcolorbox}
\vspace{-4mm}
\caption{\textbf{Prompt for generating PDF-style text.} 
An image is provided to GPT-4o-mini along with this prompt.}
\label{sup:prompt-pdfstyle}
\end{table*}

\begin{table*}[t]
\begin{tcolorbox}[colback=gray!10, colframe=black, rounded corners]
You are an AI visual assistant, and you are seeing a single image. What you see are provided within several sentences, describing the same image you are looking at. Answer all questions as you are seeing the image.
\\\\
Design a conversation between you and a person asking about this photo. The answers should be in a tone that a visual AI assistant is seeing the image and answering the question. \\
Ask diverse questions and give corresponding answers.
\\\\
Include questions asking about the visual content of the image, including the object types, counting the objects, object actions, object locations, relative positions between objects, etc. Only include questions that have definite answers: \\
(1) one can see the content in the image that the question asks about and can answer confidently; \\
(2) one can determine confidently from the image that it is not in the image. \\
Do not ask any question that cannot be answered confidently.
\\\\
Also include complex questions that are relevant to the content in the image, for example, asking about background knowledge of the objects in the image, asking to discuss about events happening in the image, etc. Again, do not ask about uncertain details. \\
Provide detailed answers when answering complex questions. For example, give detailed examples or reasoning steps to make the content more convincing and well-organized.  You can include multiple paragraphs if necessary. 
\\\\ 
You must use Japanese all the time. \\
When creating a question, start with `質問:'.\\
When creating a response, start with `回答:'.\\
After finishing a question or response, always separate them with `\textbackslash n\textbackslash n'.
\end{tcolorbox}
\vspace{-4mm}
\caption{\textbf{Prompt to generate instruction data. }
An image and paired text are provided to GPT-4o-mini along with this prompt.}
\label{sup:prompt-inst}
\end{table*}

\begin{table*}[t]
\begin{tcolorbox}[colback=gray!10, colframe=black, rounded corners]
Given a JSON array of objects, each with a `from' and `value' field, translate only the English text inside the `value' field into Japanese. Keep the special token <image>\textbackslash in the `value' field unchanged. Do not change the overall structure of the JSON. Translate all English content in the `value' field, even if it is a single word. Output only the translated JSON data and nothing else. Make sure the output is a valid JSON array that can be parsed with json.loads(). Do not include any text before or after the JSON.
\end{tcolorbox}
\vspace{-4mm}
\caption{\textbf{Prompt for translating English datasets into Japanese datasets.}
Vision-Flan~\cite{xu2024visionflan} and Image-Textualization~\cite{pi2024imagetextualization}, subsets of LLaVA OneVision~\cite{li2024llava}, are translated into Japanese using GPT-4o-mini.
}
\label{sup:prompt-translate}
\end{table*}

\section{Qualitative Analysis}\label{supsec:quali}
\begin{figure*}[t]
  \includegraphics[width=\linewidth]{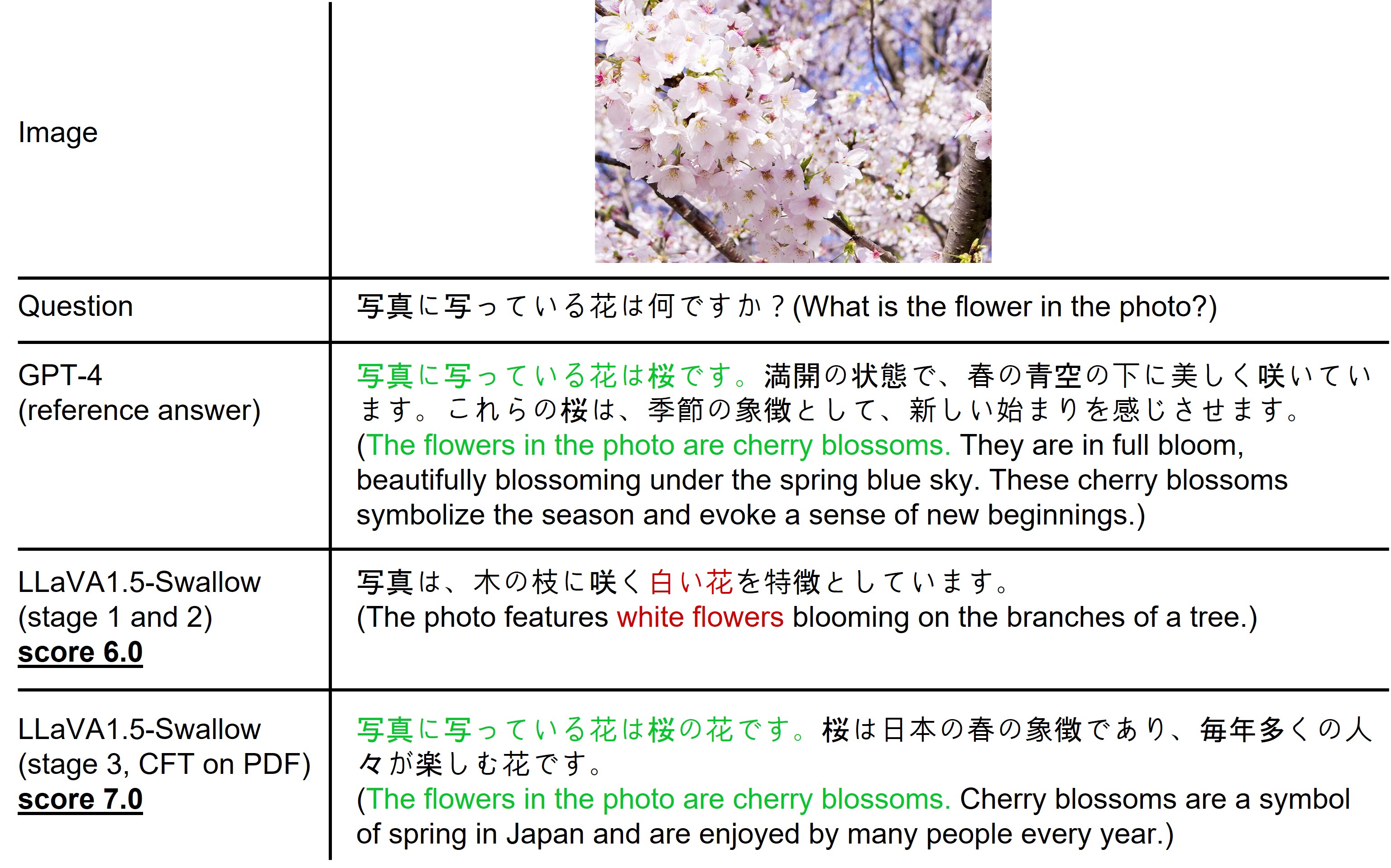}
  \vspace{-4mm}
  \caption{\textbf{Qualitative analysis on Heron-Bench.} 
  Correct parts of the responses are highlighted in green, while incorrect parts are marked in red.}
  \label{fig:heronqa1}
\end{figure*}

\begin{figure*}[t]
  \includegraphics[width=\linewidth]{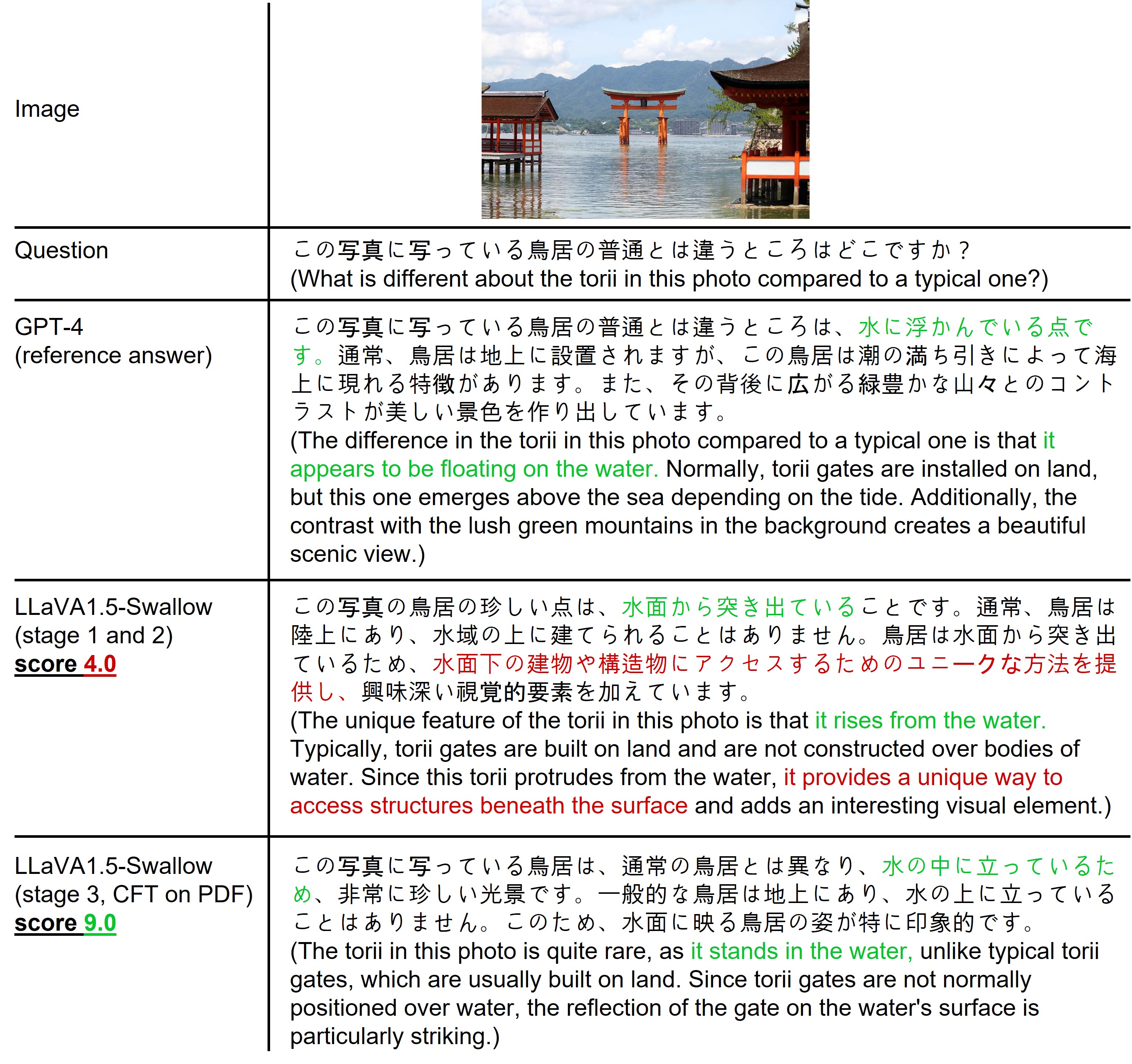}
  \vspace{-4mm}
  \caption{\textbf{Another example of qualitative analysis on Heron-Bench.}}
  \label{fig:heronqa2}
\end{figure*}

\begin{figure*}[t]
  \includegraphics[width=\linewidth]{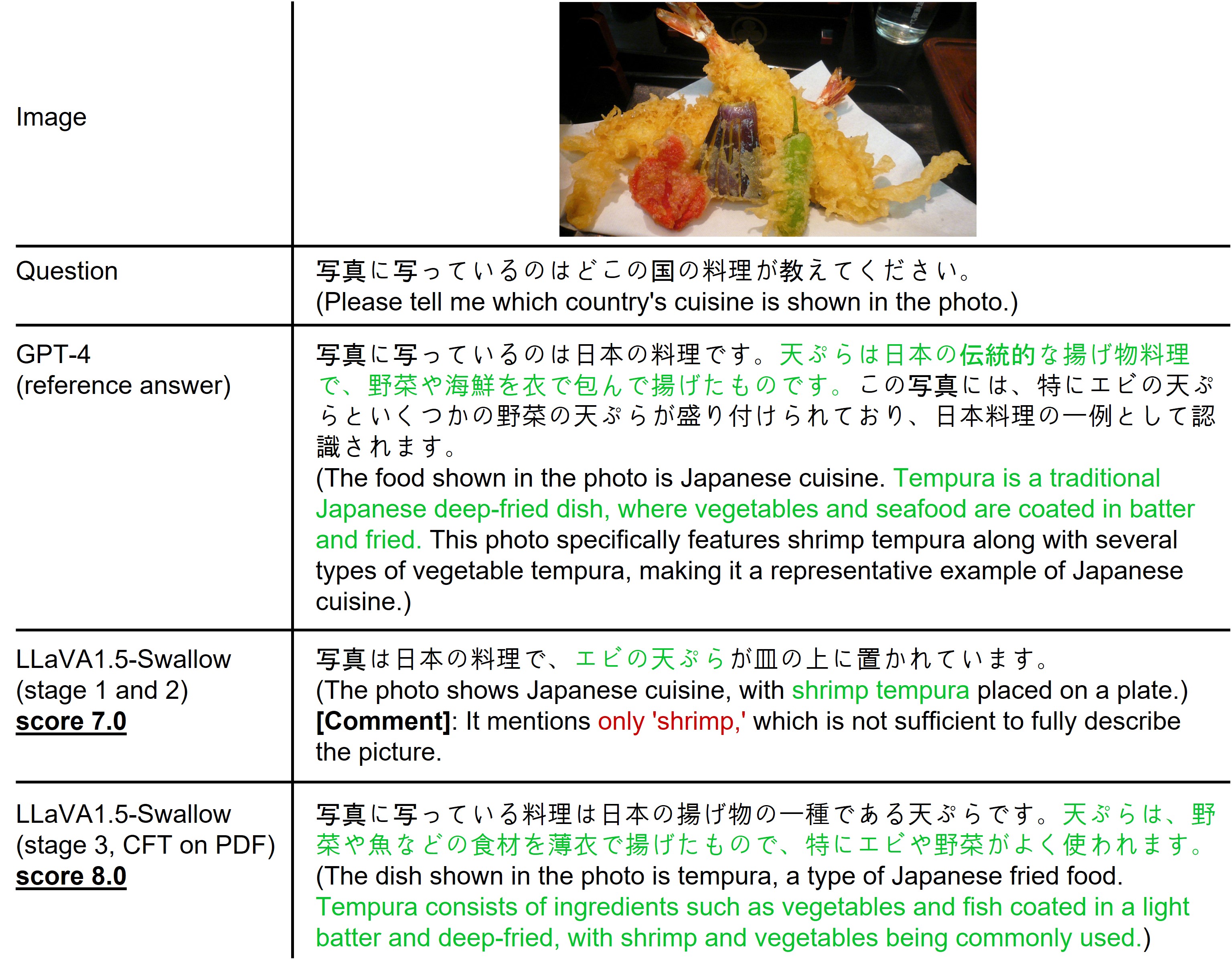}
  \vspace{-4mm}
  \caption{\textbf{Further qualitative analysis on Heron-Bench.}}
  \label{fig:heronqa3}
\end{figure*}

Figures~\ref{fig:heronqa1}, \ref{fig:heronqa2}, and \ref{fig:heronqa3} present qualitative analyses. Each figure includes an image, a question, the reference answer from GPT-4, the response from LLaVA1.5-Swallow trained up to stages 1 and 2, and the response from LLaVA1.5-Swallow further trained with stage 3 (CFT on PDF). The results show that performance improves after training up to stage 3, demonstrating the effectiveness of CFT using PDF-derived data.

\end{document}